\documentclass[sigconf, nonacm]{acmart}
\usepackage{algorithm2e}
\usepackage{xcolor}
\usepackage{graphicx}

\newcommand\vldbdoi{XX.XX/XXX.XX}

\newcommand\vldbavailabilityurl{URL_TO_YOUR_ARTIFACTS}

\usepackage{todonotes}

\begin{document}
\title{VISION: Toward a Standardized Process for Radiology Image Management at the National Level}

\author{Kathryn Knight}
\orcid{0000-0003-2976-0049}
\affiliation{%
  \institution{Oak Ridge National Laboratory}
}
\email{knightke@ornl.gov}

\author{Ioana Danciu}
\orcid{}
\affiliation{%
  \institution{Oak Ridge National Laboratory}
}
\email{danciui@ornl.gov}

\author{Olga Ovchinnikova}
\orcid{}
\affiliation{%
  \institution{Oak Ridge National Laboratory}
  \institution{Now at Thermo Fisher Scientific}
}
\email{olga.ovchinnikova@thermofisher.com}

\author{Jacob Hinkle}
\orcid{}
\affiliation{%
  \institution{Oak Ridge National Laboratory}
  \institution{Now at NVIDIA}
}
\email{jacobhinkle@gmail.com}

\author{Mayanka Chandra Shekar}
\affiliation{%
  \institution{Oak Ridge National Laboratory}
}
\email{chandrashekm@ornl.gov}

\author{Debangshu Mukherjee}
\affiliation{%
  \institution{Oak Ridge National Laboratory}
}
\email{mukherjeed@ornl.gov}

\author{Eileen McAllister}
\affiliation{%
  \institution{Oak Ridge National Laboratory}
}
\email{mcallisterea@ornl.gov}

\author{Caitlin Rizy}
\orcid{}
\affiliation{%
  \institution{Oak Ridge National Laboratory}
}
\email{rizyce@ornl.gov}

\author{Kelly Cho}
\affiliation{%
  \institution{US Department of Veterans Affairs}
}

\email{kelly.cho@va.gov}

\author{Amy C. Justice}
\affiliation{%
  \institution{US Department of Veterans Affairs}
}
\email{amy.justice2@va.gov}

\author{Joseph Erdos}
\affiliation{%
  \institution{US Department of Veterans Affairs}
}
\email{joseph.erdosmd@va.gov}

\author{Peter Kuzmak}
\affiliation{%
  \institution{US Department of Veterans Affairs}
}
\email{peter.kuzmak@va.gov}

\author{Lauren Costa}
\affiliation{%
  \institution{US Department of Veterans Affairs}
}
\email{lauren.costa@va.gov}

\author{Yuk-Lam Ho}
\affiliation{%
  \institution{US Department of Veterans Affairs}
}
\email{yuk-lam.ho@va.gov}

\author{Reddy Madipadga}
\affiliation{%
    \institution{US Department of Veterans Affairs}
}
\email{reddy.madipadga@va.gov}

\author{Suzanne Tamang}
\affiliation{%
  \institution{US Department of Veterans Affairs}
}
\email{suzanne.tamang@va.gov}

\author{Ian Goethert}
\orcid{0000-0002-1825-0097}
\affiliation{%
  \institution{Oak Ridge National Laboratory}
}
\email{goethertid@ornl.gov}

\begin{abstract}
The compilation and analysis of radiological images poses numerous challenges for researchers. The sheer volume of data as well as the computational needs of algorithms capable of operating on images are extensive. Additionally, the assembly of these images alone is difficult, as these exams may differ widely in terms of clinical context, structured annotation available for model training, modality, and patient identifiers. In this paper, we describe our experiences and challenges in establishing a trusted collection of radiology images linked to the United States Department of Veterans Affairs (VA) electronic health record database. We also discuss implications in making this repository research-ready for medical investigators. Key insights include uncovering the specific procedures required for transferring images from a clinical to a research-ready environment, as well as roadblocks and bottlenecks in this process that may hinder future efforts at automation. 
\end{abstract}

\maketitle

\begingroup

\ifdefempty{\vldbavailabilityurl}{}{
\vspace{.3cm}
}

\section{Introduction}

Since 2016, Oak Ridge National Laboratory (ORNL) has partnered with the US Department of Veteran Affairs (VA) to create a large precision medicine platform that hosts the VA’s electronic health records (EHR), a huge data set containing the medical information of over 24 million US veterans. This year, our data team piloted a program to add chest x-ray images to its medical data warehouse, enriching a repository consisting of structured data (laboratory, billing codes, medications, etc.) and unstructured clinical notes. Based on immediate successes, we expanded the pilot to include magnetic resonance images (MRI). These nontrivial tasks entailed both the secure transfer of these images as well as the processing of the image data.  After mitigating curation and access delays, the data team created an automated pipeline for transferring images and associated metadata while ensuring the security, privacy, and integrity of the image data. The project, titled ``Versatile Infrastructure System for Image Organization and aNalysis'', or VISION, ran for approximately one year from 2022 to 2023. 

The number of images we managed was noteworthy. By the end of the pilot, we received 263,000 chest x-rays and 729,000 MRI files for a sum total of 1,011,000 medical image files requiring transfer, processing, and storage. Each of these images had to be linked to the larger VA corporate data warehouse (CDW), also containing 24.8 TB of structured data and 13.7 TB of unstructured clinical notes. This paper presents our lessons learned in not just transferring these images (roughly 5 TB total), but in properly setting up a trusted and robust research environment for VA and ORNL researchers to easily query over 1 million medical image files connected to the vast amount of structured and unstructured data within the VA CDW.

The purpose of this pilot was to understand how the source repositories were organized on the clinical side, allowing for imaging data to be identified and linked to the correct patient records so they can be transferred to the research data store. There, we cleaned and organized the imaging repository into a catalogued \cite{knight_standardized_2020}, researcher-friendly, large-scale, ``big data'' effort with multiple research centers and modalities. 

Medical images pose numerous challenges related to data gathering, parsing, and quality assurance. For instance, Magudia, Bridge, Andriole, and Rosenthal \cite{magudia2021} found numerous roadblocks during their study of gathering CT scans for a machine learning study: cohort identification, retrieval, and storage were all significant roadblocks for their team. Most clinical environments (e.g., hospitals) store images in a separate location from their document-based data (patient demographics, prescription information, and so forth). Magudia, Bridge, Andriole, and Rosenthal's study is an excellent example of why this is an issue: since their team required more images than were available in one system, they had to query multiple hospital databases, where they found that retrieved exams were either mislabeled or difficult to characterize, as different systems handled exams in their own unique way (labels, identifiers, and application of administrative policies). Identifier inconsistencies result in difficulties with matching initial query data to the eventual radiology scan, since medical record numbers and accession numbers may reflect organizational policy changes that are applied inconsistently from one system to another (e.g., prepending identifiers with additional letters or numbers, or data migration issues that sometimes ``break'' links between an image and its associated documents).

\subsection{Image Transfer Pilot Goals}

As mentioned, this study sought to understand the specifics behind search and retrieval of medical images on the clinical side (where images are stored in sundry locations and tied to reporting and other patient data in a variety of ways) as well as how to effectively transfer these images into a hosted research environment, such as the one ORNL has set up for the VA. Within the research environment, this particular transfer process facilitated researcher capabilities with creating multimodal predictive modelling using deep learning (DL) techniques on chest X-Rays and MRI data. For this pilot, the iterative transfer approach (described below) enabled the effective development of DL models, where imaging data served as model input and clinical structured data from the CDW served as outcomes. Validated image batches were systematically used by researchers at ORNL to test the DL models as well as data viability prior to building models, a process which is compute-intensive when using a complete dataset.

\section{Radiology Imaging: Background}

Radiology data is a combination of text and image data, both of which are bundled together in an international standard data model, DICOM (Digital Imaging and Communications in Medicine)\footnote{International Organization for Standardization (ISO) 12052}. DICOM data is grouped into data sets, where images are bundled together with identifiers, other embedded tags, and attributes. As such, DICOM images are extremely sensitive, since patient identifiers, names, and other personally identifiable information is embedded in the image file. To aid researchers interested in radiology images (e.g., develop algorithms for automatic identification of cancers or production of radiology reports), several data sets de-identified of patient information are available to researchers. One used most heavily by the VA research team is the Medical Imaging and Data Resource Center \cite{Baughan2022}. This is a multi-institutional effort between the American College of Radiology, Radiological Society of North America, and the American Association of Physicists in Medicine. Together, they have created a publicly available image repository to assist with machine learning research in medical imaging, where de-identified images and metadata are openly accessible, minus a small percentage reserved to act as the testing set for algorithm evaluations. Other notable public data sets of chest x-rays (CXR) include ChestX-ray8, ChestX-ray14, Padchest, MIMIC-CXR, CheXpert, and VinDr-CXR. 

The ChestX-Ray8 data set is one of the first large publicly available chest x-ray data sets. which is still frequently used and widely accessible for medical imaging studies. It includes over 100,000 de-identified frontal view x-ray images of over 30,000 unique patients, all acquired from NIH Clinical Center routine care. Data includes images covering 8 different respiratory diseases: atelectasis, infiltration, pneumothorax, masses, effusion, pneumonia, cardiomegaly, and nodules \cite{Wang20173462}. The data quality is somewhat lacking, as the original radiology reports are not available, disease bounding boxes are limited, and image labels are created using natural language processing, so 100 percent accuracy of labels is not guaranteed. ChestX-ray14, an extended version of ChestX-ray8, was also recently released by the US National Institutes of Health (NIH), containing over 112,000 CXR scans from more than 30,000 patients. 

PADCHEST \cite{Bustos_2020} includes more than 160,000 images from 67,000 patients at Hospital San Juan Hospital (Spain) from 2009 to 2017. The images cover six different position views, including additional information on image acquisition and patient demography. The labels of this data set are notably rich, including 174 findings, 19 diagnoses, and 104 localizations (as compared to approximately 14 binary labels for each of the other data sets). 

MIMIC-CXR \cite{mimiccxr} is a data set slightly larger than ChestX-Ray8 (around 377,000 images) that also includes radiology report text along with the chest radiographs. In addition to the images, 227,835 radiographic studies are also included along with deidentified DICOMs. This is the data set used most often by ORNL medical image researchers, and links to the MIMIC-IV EHR data set. CheXpert is another data set that is similar in size to MIMIC-CXR (around 224,000 images). While it does not link to a full EHR data set, CheXpert does have high-quality labels including more ground truths from radiologists than the other available data sets. 

Finally, the VinDr-CXR \cite{vindrcxr}, is a collection of 18,000 images collected at Hospital 108 and Hanoi Medical University Hospital, two large hospitals in Vietnam. Though smaller than the other data sets, this set is of high quality: the published data set consists of 18,000 postero-anterior (PA) view CXR scans that come with both the localization of critical findings and the classification of common thoracic diseases. Every image in the training set (15000 images) was labelled by 3 radiologists, and for the 3000 test images, 5 radiologists read each one. These have been labeled with 22 critical findings (local labels) and 6 diagnoses (global labels), with each finding indicated by a bounding box.

\subsection{Medical Imaging Data Storage}

Most medical data are generated by the electronic healthcare record system and ancillary systems such as pathology, pharmacy, radiology, etc . All of these entities are relatively interoperable in that the data are linked by common identifiers, such as patient or encounter IDs. Clinical/corporate Data Warehouses (CDW) serve to aggregate these various data sources and enable a variety of clinical research, such as phenomic analyses. Multiple resources of medical data are available to the medical community. i2b2, developed by Harvard Medical School and funded by the National Institute of Health, integrates clinical and genomic data into a data warehouse-style star schema. Another widely-used warehouse solution is Dr. Warehouse, developed at Necker's Children's Hospital in Paris, which is a framework that integrates 26 data sources (EHR, biological data, and other document-based data) from over 500,000 patients and enables multiple views for clinicians to generate medical narrative reports\cite{allofus}. Other proposed solutions include Stanford's STRIDE\cite{stride}, radBank\cite{Rubin2008210} and radTF\cite{do20102039}, Vanderbilt's research data warehouse solution\cite{DANCIU201428}, Mayo Clinic's Enterprise Data Trust\cite{Chute2010131}, DW4TR\cite{dw4tr}, METEOR\cite{meteor}, and the SMart EYeE DATabase\cite{smarteye}.

To make data aggregation and therefore analyses easier, several efforts have emphasized creating standards for medical data archiving. The OHDSI initiative has published a common data model, the OMOP-CDM (Observational Medical Outcomes Partnership). This contributed to standardization for medical data warehousing, and has been adopted by many partners over the world. i2b2 is OMOP-compatible, as is Dr. Warehouse. However, OMOP is specific for clinical data, needs additional standards to accommodate the metadata associated with medical images.  

Since medical imaging data have large storage requirements and are not in a format easily accommodated by traditional data warehouses, they are rarely integrated into CDWs (i.e., stores of tabular data and text that are easily grouped into fields and tables). Instead, medical images are frequently stored separately and linked to any related patient data and text-based study interpretation via an identifier (or multiple identifiers, depending on the images). Of the solutions listed above, they either do not include images as part of the main data model, or the system inadequately addresses the logistical challenges of managing image data collection, integration, and data quality assurance. One promising recent development in standardization of medical imaging data is the recent publication of the Radiology Common Data Model  (R-CDM) \cite{rcdm}, though further research is necessary to understand how this data model may be integrated into current radiology image extract transform and load (ETL) workflows, such as those described in this paper. 

\section{Methods}

 This project was approved by the VA Central Institutional Review Board (IRB).

The processing solution was carefully designed, implemented, validated, and documented. Because the transfer process is labor-intensive, images were transferred in batches, where multiple imaging studies are transmitted as a group. This approach allowed for iterative development, and benefited both the persons facilitating the transfers from the clinical environment as well as the researchers.
 Figure 1 shows an overview of the transfer process; the blue box corresponds to the Clinical Environment pipeline at the VA, and the green to the Research Environment at ORNL.  

\begin{figure}
  \centering
  \includegraphics[width=\linewidth]{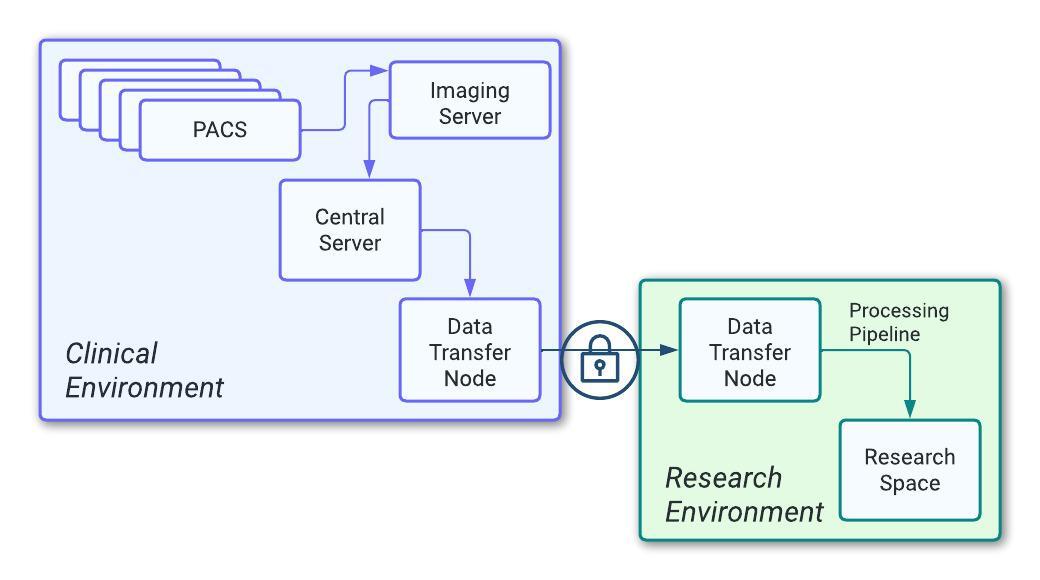}
  \caption{Overview of DICOM Image Transfer Pipeline (VA in blue and ORNL in green)}
  \label{fig:pipeline}
\end{figure}

Below are the detailed steps that each pipeline entails: one from the VA's clinical environment, where the images were searched, aggregated, and sent, and one from ORNL's research environment, where images were received, processed, and stored for use. Both pipelines require input from clinical and consulting Subject Matter Experts (SMEs).

\RestyleAlgo{ruled}

\begin{algorithm}[hbt!]
\SetAlgorithmName{Pipeline}{}{}
\caption{\color{blue}\textbf{\textit{Clinical Environment (VA)}}}
Step 1: VA and ORNL agree on a batch size for image transfer(s)

Step 2: Process SME requests batch from imaging SME

Step 3: Imaging SME identifies the radiological studies that meet the current batch size

Step 4: This list of studies (accession numbers) is requested from the facility administrator

Step 5: The facility extracts the images based on the study list and notifies the transfer team of the staging location

Step 6: The transfer team transfers the files to ORNL and notifies the VA SME

Step 7: VA SME notifies ORNL of successful transmission

Step 8: Upon receipt confirmation from ORNL, transfer team deletes batch from the staging location
\end{algorithm}

The clinical environment pipeline's first four steps require input from a minimum of three individuals. Step 1 requires collaborative decision making between the clinical and research sides, where a designated batch size (i.e., sum total size or number of files) is agreed upon. Steps 2 and 3 entail collaboration between a Process SME and Imaging SME, who provide a facility administrator with a list of accession numbers (step 4) to extract the images from the VA's research environment, a large facility data warehouse (step 5). Steps 6 - 8 entail the actual image transfer, which also requires direct communication between individuals to confirm image receipt and subsequent deletion from the staging area in the clinical environment. 

\RestyleAlgo{ruled}

\begin{algorithm}[hbt!]
\SetAlgorithmName{Pipeline}{}{}
\caption{\color{teal}\textbf{\textit{Research Environment (ORNL)}}}
Step 1: Batch receipt confirmed by ORNL IT SME

Step 2: File hashes created and stored as metadata by ORNL Engineering SME

Step 3: Images scanned for file and study related metadata and persisted in research workspace by ORNL Engineering SME

Step 4: Report text and demographic data persisted in the research workspace

Step 5: Data profiled for quality and reporting

\end{algorithm}

The research environment's pipeline begins with confirmed receipt of the images, Step 7 in the clinical environment pipeline. File hashes are generated and saved as a ``snapshot'' of what has been received. Next are scripts that open each image to extract key DICOM header metadata elements as well as information about the file itself, such as size and location, which is saved to the research workspace. Using these metadata, we then extract related structured and unstructured data from the larger VA data warehouse. This includes demographic information, study dates, and report texts that can be linked directly to the image data. Having all these data in the research workspace minimizes the need for researchers to perform time-consuming data engineering tasks. Finally, this process profiles the resulting data on a dashboard that generates reports (e.g., any metadata errors on read) for both operational and research use. 

\subsection{Data Collection}

The chest x-ray pilot began by using images from a single VA site to limit the complexity of data collection. Even still, each site can contain multiple Picture Archiving and Communicating System (PACS) interfaces. To ensure close collaboration between ORNL and the VA for the duration of the project, we chose a VA site that included a VA imaging subject matter expert to assist with consulting and validation. Furthermore, as this pilot was to develop an entirely new ETL process for medical images, we chose an iterative approach to ensure continuous process improvement. 

\subsubsection{Image Identification}
First, image studies are identified by a subject matter expert at the VA. To do this, they need to identify the accession numbers associated with each image (the DICOM standard uses a system-generated accession number to link the image study order to the imaging system). These associated accession numbers are divided into manageable batches and then used by the VA SME to pull the image files. Image extraction was noticeably slower during regular business hours, and overall the process was relatively slow. For instance, 40 accession numbers took 9 hours to extract from the system. A different, smaller batch of 10 took somewhat less time, but also required human intervention and did not add to overall productivity of the process. Furthermore, running a job on more than one server required special care to ensure the correct data was pulled. Overall, this is crucial information for any effort at scaling, as any move toward automation will need to consider business hours, system overall capability, time required for overall image extraction, as well as process quality control. 

After extraction, the image files are transferred to ORNL, after which the accession numbers are then used at ORNL to validate the image transfers. File transfer time was also time-dependent, as the image SME noticed that file transfer to ORNL took significantly less time in the morning than the afternoon. Again, as efforts toward automation of this process grow in size and scope, these details will be essential to note and understand to streamline the overall process. 

\subsubsection{Batch Transfers}
The ORNL copy of VA CDW receives daily updates from the VA using a secure connection between the sites. An Extract, Transform, and Load (ETL) process purpose built by the VA keeps the ORNL copy of CDW current. Since this process is relied upon for other research, it was imperative that we do not interfere with that process. The batch approach to transfers allows us to the potential for saturating the connection and interfering with the existing ETL process. By limiting the number of studies in a batch, we were able to control the load on the transfer pipeline based on transfer size and time of day.

\subsubsection{Processing Pipeline and Data Integration}

Upon arrival to the data transfer node, an automated process creates a hash value of each image file to ensure image integrity. This hash value allows for easy validation of images by researchers, e.g., that the image hasn't been subsequently corrupted or modified after being transferred and stored on ORNL's infrastructure. Once released to the research space, the processing pipeline performs a number of critical steps which is the final step in the data collection process. 

First, the processing pipeline reads selected DICOM header information for each transferred image, which validates that the file transferred is truly a DICOM image and not another image format. This is a crucial step, as during our pilot, we found a small number of images that failed. However, on closer examination, we found that a legacy system stored the image and header data separately. When these studies were converted to the modern DICOM standard, the four byte "DICM" prefix was omitted--so the failed images were, in fact, DICOM images. As there are plans to eventually automate this process, it is essential that we understand such discrepancies and idiosyncrasies, to help truly distinguish between undesired data and data arriving in unexpected formats due to legacy systems. The team uses an open source Python library, pydicom\cite{pydicom}, to read in the DICOM header metadata, which has the ability to ``force'' read the header information, allowing us to capture that status of the image as modern, legacy, or corrupt. 

Next, the information extracted from the DICOM header is persisted in a database set aside specifically for the research project. Researchers can access the selected DICOM elements quickly and easily without having to go back to the actual image files. This includes device related information such as kilovoltage peak (kVp), exposure time, and resolution as well as manufacturer and software version. It also includes patient-related information such as procedure description, image type, and view position. This also includes metadata information such as accession number, file location, and size. The accession number can be validated against the accession number list provided by the VA during the first part of the data collection process. 

Next, the pipeline is used to snapshot and persist key clinical and demographic information from the VA's Corporate Data Warehouse (CDW). Having this information readily available alongside the image data allows researchers to minimize time spent on data engineering and maximize time spent on research. 

For now, the entire collection and integration pipeline runs upon arrival of a new batch of images. As new elements in the DICOM standard and/or in the clinical warehouse are identified, they are automatically collected in the next batch ingress. The  pipeline can process approximately 157 images per second (500k+ files per hour). Future versions of the pipeline will include hardware-accelerated processing and ultimately will move to a differential approach for processing only newly arrived data. Other than a small number of metadata elements (e.g., batch number), the pipeline is fully contained and thus could be easily automated.

Finally, we profiled and pushed the metadata to an interactive dashboard that is useful for operational status as well as research exploration. While the primary goal of the pilot was not to generate a list of standardized data terms and processes related to image data collection (e.g., formats, roles of people involved, processes, platforms, governing entities, and so on), this is still something the team views as essential to support research related to electronic health records. 

\section{Challenges and Insights}

The challenges related to medical image integration with clinical data cannot be understated. The research environment hosted at ORNL is a 38.5 TB data warehouse containing over 68 schemas and over 1,600 tables holding over 20 years of patient data for US Veterans. Access to data within is based on user status restrictions with multiple granularity levels, so integrating image data to this vast data store requires attention to multiple data components. Below we will describe some of the key lessons learned during this pilot, with an eye to how this may scale when additional modalities and/or facilities are added to the process. 

One standout challenge was related to data quality and trust, which will be described in detail below. Bad labels and inconsistent identifiers may require months of necessary data reconciliation and cleanup work before an investigation can even begin. Thus, eliminating the task of linking data and potentially dealing with incorrect or otherwise corrupted identifiers across data stores, coupled with highly performant storage, allows researchers to focus solely on scientific research rather than time-consuming data wrangling and quality control. 

\subsection{Summary of Lessons Learned}

There are many key insights this pilot project brought to light. First, metadata management is not only a huge problem, but the systems designed to manage transferring medical images will need to address this issue up front, not as an afterthought. ORNL spent many hours clarifying what the clinical side intended to send versus what was actually received, much of which was due to issues with metadata such as accession number formatting, duplication, or other header data within the image files that required manual scrutiny. Some of this was due to legacy data, which is also a reality that any efforts toward automation will need to address: legacy data may reflect former standards and policy applications (e.g., prepending identifiers with additional characters) that can hinder data processing pipelines. Without properly addressing this, there is little chance of effectively automating this process, which will subsequently affect scalability. 

Furthermore, it is essential that image search, retrieval, and transfer do not interfere with existing business processes or systems tied to day-to-day clinical care. Many hospital systems are in use during regular hours by physicians, and cannot be ``tied up'' by extensive search/retrieval tasks.

We learned a number of lessons from the process itself. While developing the process iteratively did allow the team to add features to the pipeline quickly, it also significantly added to the amount of technical debt accrued during the QA process. For example, during the pilot we decided to include accession and file metadata with each batch in the form of manifest files. These manifest files can be used to confirm that all intended files were transferred. However, these manifest files do not exist for previously transferred batches. Therefore, automated processes need account for this absence without marking the batch as incomplete. We were also able to understand how many manual processes are involved in image search, aggregation, and transfer, and have a much clearer conception of what can be investigated for streamlining. Visualization of descriptive statistics was useful by operations and planning teams for such activities as understanding batch size and disk usage. Researchers also found this useful for understanding distributions based on elements such a manufacturer and modality.  

Ultimately, this pilot indicated that there is a real research need for having images linked to both structured and unstructured clinical data within the same environment. Clinical systems are not set up for research (e.g., connected to HPC or systems that enable aggregation and analysis of data), so moving these images to a location where researchers have immediate and easy access to HPC, ample storage, and software packages is essential. Our investigation not only highlights this need, but also underscores what specific challenges should be addressed when developing a scalable system that not only transfers medical images of various sizes and modalities, but also combines them with other complex clinical data. 

\subsection{Data quality and trust}
The pilot identified potential issues with data quality in what ultimately became two stages. The first stage sought to understand image data at a very high level: images were inventoried based on expected file count. Meaning, a basic chest x-ray study can consist of minimum two images; a front view and a side view, plus four to six modified versions of the front and side view, each captured by the same hardware in relevant angles by radiologist. Thus, if 10 x-ray studies are anticipated, then the approximate total expected image files will be around 80. As a result, during the first stage the team was able to anticipate an approximate average of 7-8 files per study, and gauge the expected file ingest volume accordingly.

However, once we added MRI studies to the data ingest pipeline, the team quickly learned that such a simple, high-level approach would not scale for more complex image data and research goals. This is because a single MRI image can consist of a wide number of images, generally between 200 and 1,600 image files. Thus, we determined that there was a second stage necessary to understand how to account for the files we expected from the VA. 

The second stage, then, became an iterative process. The team developed and incorporated a procedure to include basic metadata with each batch of files in the form of two text files that are generated during the file transfer. These files contain study and file metadata, respectively. As the format of these files stabilized, the ingest data pipeline parsed and tested each batch. This allowed for more advanced features such as rejecting a batch while automating what is currently still a manually labor intensive process. This second stage has revealed a number of opportunities for improvement, as the complexity of these file transfers cannot be understated. Potential issues with ensuring data quality include proper management of duplicate files, studies, and/or batches, as well as duplicated accession numbers and other identifiers. 

As stewards of the clinical data, it is imperative that the research team not just successfully receive and store the images, but also account for what has been received. This accounting allowed the team to effectively communicate to the clinical research team if data were missing, received in duplicate, or if data that should not have been sent was accidentally communicated. This was essential to keep the team compliant with any IRB restrictions as well as communicate to end users whether accounting for duplicate or missing information was necessary (and how they should do that).

\subsection{Access, Compute, and Cost}

Several areas were not addressed with this pilot project. Logistics surrounding secure access to the images (on both the clinical and research sides) and other necessary security measures are built upon security and privacy policies already in place as set by the VA and managed by ORNL's Knowledge Discovery Infrastructure (KDI) team. Furthermore, the research environment within the ORNL enclave uses a Lakehouse approach\cite{begoli2021} to access control and security. Other investigations and subsequent systems development may need to incorporate policies and procedures related to IRB restrictions or other access and security concerns. 

Furthermore, the ORNL environment is already set up for medical analyses on high performance machines, all with access to ample file storage. Logistics and cost related to standing up appropriate hardware as well as staff to manage HPC systems was out of scope of this pilot. For the next iteration of our investigation regarding medical image transfer and storage, the team intends to focus not just on scalability, but on cost, and what cost reduction measures may be taken as this effort scales. 

\section{Conclusion}

Healthcare imaging systems pose numerous challenges for researchers wishing to conduct large imaging studies. Having a trusted paradigm for gathering and organizing imaging in a research environment allows researchers to focus primarily on research and less on planning and engineering. With this pilot, we sought to understand what this paradigm entails. Of significant importance is the quality of the image metadata, which may vary depending on legacy systems and/or how policy choices are reflected in what is recorded in the metadata. Additionally, understanding the extent to which human interaction was necessary for the transfer is key, as any system in place to enable automated image transfer will need to account for clinic-specific workflows, that, when scaled to multiple institutions and/or modalities (that may involve additional systems, departments, policies, and so on), will grow exponentially complex.

Our investigation was initially based on the data and workflows for one clinic, and one image modality only (chest x-rays) and later expanded to abdominal MRIs. As such, the problems we uncovered with 1 million images and 5 terabytes of data will grow exponentially when scaled to cover all possible medical image modalities as well as all of the clinical facilities that store these images. Estimating the sum total data this will entail (along with the variety of institutional systems and policies affecting the data quality) is outside the scope of this project, and part of why we chose to approach this problem by conducting a study of just one location and modality. As machine learning research on medical images increases, however, understanding the processes involved in ingressing and storing these images is essential for developing large research environments with access to a data commons.

\begin{acks}
Notice: This manuscript has been authored by UT-Battelle, LLC under Contract No. DE-AC05-00OR22725 with the U.S. Department of Energy. The publisher, by accepting the article for publication, acknowledges that the U.S. Government retains a non-exclusive, paid up, irrevocable, world-wide license to publish or reproduce the published form of the manuscript, or allow others to do so, for U.S. Government purposes. The DOE will provide public access to these results in accordance with the DOE Public Access Plan (http://energy.gov/downloads/doe-public-access-plan).

This research is supported by the Million Veteran Program, Office of Research and Development, Veterans Health Administration, and was supported by award MVP-000. This publication does not represent the views of the Department of Veteran Affairs or the United States Government.

The VISION project team would like to acknowledge the support of the Knowledge Discovery Infrastructure (KDI) team, Hope Cook, and Dallas Sacca at Oak Ridge National Laboratory. Their assistance and technical expertise were integral to the success of this project.
\end{acks}


\bibliographystyle{ACM-Reference-Format}
\bibliography{main}


\begin{thebibliography}{19}


\ifx \showCODEN    \undefined \def \showCODEN     #1{\unskip}     \fi
\ifx \showDOI      \undefined \def \showDOI       #1{#1}\fi
\ifx \showISBNx    \undefined \def \showISBNx     #1{\unskip}     \fi
\ifx \showISBNxiii \undefined \def \showISBNxiii  #1{\unskip}     \fi
\ifx \showISSN     \undefined \def \showISSN      #1{\unskip}     \fi
\ifx \showLCCN     \undefined \def \showLCCN      #1{\unskip}     \fi
\ifx \shownote     \undefined \def \shownote      #1{#1}          \fi
\ifx \showarticletitle \undefined \def \showarticletitle #1{#1}   \fi
\ifx \showURL      \undefined \def \showURL       {\relax}        \fi
\providecommand\bibfield[2]{#2}
\providecommand\bibinfo[2]{#2}
\providecommand\natexlab[1]{#1}
\providecommand\showeprint[2][]{arXiv:#2}

\bibitem[\protect\citeauthoryear{??}{str}{2009}]%
        {stride}
 \bibinfo{year}{2009}\natexlab{}.
\newblock \showarticletitle{STRIDE--An integrated standards-based translational
  research informatics platform}. In \bibinfo{booktitle}{\emph{AMIA Annual
  Symposium Proceedings}}.
\newblock


\bibitem[\protect\citeauthoryear{Baughan, Whitney, Drukker, Sahiner, Hu, Kim
  Grace~Hyun, McNitt-Gray, Myers, and Giger}{Baughan et~al\mbox{.}}{2022}]%
        {Baughan2022}
\bibfield{author}{\bibinfo{person}{Natalie Baughan},
  \bibinfo{person}{Heather~M. Whitney}, \bibinfo{person}{Karen Drukker},
  \bibinfo{person}{Berkman Sahiner}, \bibinfo{person}{Tingting Hu},
  \bibinfo{person}{J. Kim Grace~Hyun}, \bibinfo{person}{Michael McNitt-Gray},
  \bibinfo{person}{Kyle Myers}, {and} \bibinfo{person}{Maryellen~L. Giger}.}
  \bibinfo{year}{2022}\natexlab{}.
\newblock \showarticletitle{Sequestration of Imaging Studies in MIDRC: A
  Multi-Institutional Data Commons}.
\newblock \bibinfo{journal}{\emph{Progress in Biomedical Optics and Imaging -
  Proceedings of SPIE}}  \bibinfo{volume}{12035}.
\newblock
\urldef\tempurl%
\url{https://doi.org/10.1117/12.2610239}
\showDOI{\tempurl}


\bibitem[\protect\citeauthoryear{Begoli, Goethert, and Knight}{Begoli
  et~al\mbox{.}}{2021}]%
        {begoli2021}
\bibfield{author}{\bibinfo{person}{Edmon Begoli}, \bibinfo{person}{Ian
  Goethert}, {and} \bibinfo{person}{Kathryn Knight}.}
  \bibinfo{year}{2021}\natexlab{}.
\newblock \showarticletitle{A Lakehouse Architecture for the Management and
  Analysis of Heterogeneous Data for Biomedical Research and Mega-biobanks}. In
  \bibinfo{booktitle}{\emph{2021 IEEE International Conference on Big Data (Big
  Data}}. \bibinfo{publisher}{IEEE}, \bibinfo{pages}{4643--4651}.
\newblock
\urldef\tempurl%
\url{https://doi.org/10.1109/BigData52589.2021.9671534}
\showDOI{\tempurl}


\bibitem[\protect\citeauthoryear{Bustos, Pertusa, Salinas, and de~la
  Iglesia-Vay{\'{a}}}{Bustos et~al\mbox{.}}{2020}]%
        {Bustos_2020}
\bibfield{author}{\bibinfo{person}{Aurelia Bustos}, \bibinfo{person}{Antonio
  Pertusa}, \bibinfo{person}{Jose-Maria Salinas}, {and} \bibinfo{person}{Maria
  de~la Iglesia-Vay{\'{a}}}.} \bibinfo{year}{2020}\natexlab{}.
\newblock \showarticletitle{PadChest: A large chest x-ray image dataset with
  multi-label annotated reports}.
\newblock \bibinfo{journal}{\emph{Medical Image Analysis}}
  \bibinfo{volume}{66} (\bibinfo{date}{dec} \bibinfo{year}{2020}),
  \bibinfo{pages}{101797}.
\newblock
\urldef\tempurl%
\url{https://doi.org/10.1016/j.media.2020.101797}
\showDOI{\tempurl}


\bibitem[\protect\citeauthoryear{Chute, Beck, Fisk, and Mohr}{Chute
  et~al\mbox{.}}{2010}]%
        {Chute2010131}
\bibfield{author}{\bibinfo{person}{Christopher~G Chute},
  \bibinfo{person}{Scott~A Beck}, \bibinfo{person}{Thomas~B Fisk}, {and}
  \bibinfo{person}{David~N Mohr}.} \bibinfo{year}{2010}\natexlab{}.
\newblock \showarticletitle{The Enterprise Data Trust at Mayo Clinic: A
  semantically integrated warehouse of biomedical data}.
\newblock \bibinfo{journal}{\emph{Journal of the American Medical Informatics
  Association}} \bibinfo{volume}{17}, \bibinfo{number}{2}
  (\bibinfo{year}{2010}), \bibinfo{pages}{131 – 135}.
\newblock


\bibitem[\protect\citeauthoryear{Danciu, Cowan, Basford, Wang, Saip, Osgood,
  Shirey-Rice, Kirby, and Harris}{Danciu et~al\mbox{.}}{2014}]%
        {DANCIU201428}
\bibfield{author}{\bibinfo{person}{Ioana Danciu}, \bibinfo{person}{James~D.
  Cowan}, \bibinfo{person}{Melissa Basford}, \bibinfo{person}{Xiaoming Wang},
  \bibinfo{person}{Alexander Saip}, \bibinfo{person}{Susan Osgood},
  \bibinfo{person}{Jana Shirey-Rice}, \bibinfo{person}{Jacqueline Kirby}, {and}
  \bibinfo{person}{Paul~A. Harris}.} \bibinfo{year}{2014}\natexlab{}.
\newblock \showarticletitle{Secondary use of clinical data: The Vanderbilt
  approach}.
\newblock \bibinfo{journal}{\emph{Journal of Biomedical Informatics}}
  \bibinfo{volume}{52} (\bibinfo{year}{2014}), \bibinfo{pages}{28--35}.
\newblock
\showISSN{1532-0464}
\urldef\tempurl%
\url{https://doi.org/10.1016/j.jbi.2014.02.003}
\showDOI{\tempurl}
\newblock
\shownote{Special Section: Methods in Clinical Research Informatics.}


\bibitem[\protect\citeauthoryear{do, Wu, Biswal, Kamaya, and Rubin}{do
  et~al\mbox{.}}{2010}]%
        {do20102039}
\bibfield{author}{\bibinfo{person}{Bao~H. do}, \bibinfo{person}{Andrew Wu},
  \bibinfo{person}{Sandip Biswal}, \bibinfo{person}{Aya Kamaya}, {and}
  \bibinfo{person}{Daniel~L. Rubin}.} \bibinfo{year}{2010}\natexlab{}.
\newblock \showarticletitle{Informatics in Radiology RADTF: A Semantic
  Search-enabled, Natural Language Processor-generated radiology teaching
  file}.
\newblock \bibinfo{journal}{\emph{Radiographics}} \bibinfo{volume}{30},
  \bibinfo{number}{7} (\bibinfo{year}{2010}), \bibinfo{pages}{2039 – 2048}.
\newblock


\bibitem[\protect\citeauthoryear{Hu}{Hu}{2011}]%
        {dw4tr}
\bibfield{author}{\bibinfo{person}{et~al. Hu, H.}}
  \bibinfo{year}{2011}\natexlab{}.
\newblock  \bibinfo{volume}{44}, \bibinfo{number}{6} (\bibinfo{year}{2011}),
  \bibinfo{pages}{1004–1019}.
\newblock
\urldef\tempurl%
\url{https://doi.org/10.1016/j.jbi.2011.08.003}
\showDOI{\tempurl}


\bibitem[\protect\citeauthoryear{Johnson, Pollard, Berkowitz, Greenbaum,
  Lungren, Deng, Mark, and Horng}{Johnson et~al\mbox{.}}{2019}]%
        {mimiccxr}
\bibfield{author}{\bibinfo{person}{Alistair E.~W. Johnson},
  \bibinfo{person}{Tom~J. Pollard}, \bibinfo{person}{Seth~J. Berkowitz},
  \bibinfo{person}{Nathaniel~R. Greenbaum}, \bibinfo{person}{Matthew~P.
  Lungren}, \bibinfo{person}{Chih-ying Deng}, \bibinfo{person}{Roger~G. Mark},
  {and} \bibinfo{person}{Steven Horng}.} \bibinfo{year}{2019}\natexlab{}.
\newblock \showarticletitle{MIMIC-CXR, a de-identified publicly available
  database of chest radiographs with free-text reports}.
\newblock \bibinfo{journal}{\emph{Scientific Data}} \bibinfo{volume}{6},
  \bibinfo{number}{1} (\bibinfo{year}{2019}), \bibinfo{pages}{317}.
\newblock
\urldef\tempurl%
\url{https://doi.org/10.1038/s41597-019-0322-0}
\showDOI{\tempurl}


\bibitem[\protect\citeauthoryear{Knight, Honerlaw, Danciu, Linares, Ho, Gagnon,
  Rush, Gaziano, Begoli, and Cho}{Knight et~al\mbox{.}}{2020}]%
        {knight_standardized_2020}
\bibfield{author}{\bibinfo{person}{Kathryn~E. Knight},
  \bibinfo{person}{Jacqueline Honerlaw}, \bibinfo{person}{Ioana Danciu},
  \bibinfo{person}{Franciel Linares}, \bibinfo{person}{Yuk-Lam Ho},
  \bibinfo{person}{David~R. Gagnon}, \bibinfo{person}{Everett Rush},
  \bibinfo{person}{J.~Michael Gaziano}, \bibinfo{person}{Edmon Begoli}, {and}
  \bibinfo{person}{Kelly Cho}.} \bibinfo{year}{2020}\natexlab{}.
\newblock \showarticletitle{Standardized {Architecture} for a {Mega}-{Biobank}
  {Phenomic} {Library}: {The} {Million} {Veteran} {Program} ({MVP})}.
\newblock \bibinfo{journal}{\emph{AMIA Jt Summits Transl Sci Proc}}
  \bibinfo{volume}{2020} (\bibinfo{date}{May} \bibinfo{year}{2020}),
  \bibinfo{pages}{326--334}.
\newblock
\showISSN{2153-4063}
\urldef\tempurl%
\url{https://www.ncbi.nlm.nih.gov/pmc/articles/PMC7233040/}
\showURL{%
\tempurl}


\bibitem[\protect\citeauthoryear{Kortüm~KU}{Kortüm~KU}{2017}]%
        {smarteye}
\bibfield{author}{\bibinfo{person}{et~al. Kortüm~KU}.}
  \bibinfo{year}{2017}\natexlab{}.
\newblock \showarticletitle{Using electronic health records to build an
  ophthalmologic data warehouse and visualize patients' data}.
\newblock \bibinfo{journal}{\emph{American Journal of Ophthalmology}}
  \bibinfo{volume}{178} (\bibinfo{year}{2017}), \bibinfo{pages}{84–93}.
\newblock
\urldef\tempurl%
\url{https://doi.org/10.1016/j.ajo.2017.03.026}
\showDOI{\tempurl}


\bibitem[\protect\citeauthoryear{Magudia, Bridge, Andriole, and
  Rosenthal}{Magudia et~al\mbox{.}}{2021}]%
        {magudia2021}
\bibfield{author}{\bibinfo{person}{K. Magudia}, \bibinfo{person}{C.~P. Bridge},
  \bibinfo{person}{K.~P. Andriole}, {and} \bibinfo{person}{M.~H. Rosenthal}.}
  \bibinfo{year}{2021}\natexlab{}.
\newblock \showarticletitle{The Trials and Tribulations of Assembling Large
  Medical Imaging Datasets for Machine Learning Applications}.
\newblock \bibinfo{journal}{\emph{Journal of digital imaging}}
  \bibinfo{volume}{34}, \bibinfo{number}{6} (\bibinfo{year}{2021}),
  \bibinfo{pages}{1424–1429}.
\newblock
\urldef\tempurl%
\url{https://doi.org/10.1007/s10278-021-00505-7}
\showDOI{\tempurl}


\bibitem[\protect\citeauthoryear{Mason}{Mason}{[n.d.]}]%
        {pydicom}
\bibfield{author}{\bibinfo{person}{et~al Mason, D.~L.}}
  \bibinfo{year}{[n.d.]}\natexlab{}.
\newblock \bibinfo{title}{pydicom: An open source DICOM library}.
\newblock
\newblock
\urldef\tempurl%
\url{https://github.com/pydicom/pydicom}
\showURL{%
\tempurl}
\newblock
\shownote{Online, accessed 2023-02-23.}


\bibitem[\protect\citeauthoryear{Nguyen, Lam, Le, Pham, Tran, Nguyen, Le, Pham,
  Tong, Dinh, Do, Doan, Nguyen, Nguyen, Nguyen, Hoang, Phan, Nguyen, Ho, Ngo,
  Nguyen, Nguyen, Dao, and Vu}{Nguyen et~al\mbox{.}}{2020}]%
        {vindrcxr}
\bibfield{author}{\bibinfo{person}{Ha~Q. Nguyen}, \bibinfo{person}{Khanh Lam},
  \bibinfo{person}{Linh~T. Le}, \bibinfo{person}{Hieu~H. Pham},
  \bibinfo{person}{Dat~Q. Tran}, \bibinfo{person}{Dung~B. Nguyen},
  \bibinfo{person}{Dung~D. Le}, \bibinfo{person}{Chi~M. Pham},
  \bibinfo{person}{Hang T.~T. Tong}, \bibinfo{person}{Diep~H. Dinh},
  \bibinfo{person}{Cuong~D. Do}, \bibinfo{person}{Luu~T. Doan},
  \bibinfo{person}{Cuong~N. Nguyen}, \bibinfo{person}{Binh~T. Nguyen},
  \bibinfo{person}{Que~V. Nguyen}, \bibinfo{person}{Au~D. Hoang},
  \bibinfo{person}{Hien~N. Phan}, \bibinfo{person}{Anh~T. Nguyen},
  \bibinfo{person}{Phuong~H. Ho}, \bibinfo{person}{Dat~T. Ngo},
  \bibinfo{person}{Nghia~T. Nguyen}, \bibinfo{person}{Nhan~T. Nguyen},
  \bibinfo{person}{Minh Dao}, {and} \bibinfo{person}{Van Vu}.}
  \bibinfo{year}{2020}\natexlab{}.
\newblock \showarticletitle{VinDr-CXR: An open dataset of chest X-rays with
  radiologist's annotations}.
\newblock  (\bibinfo{year}{2020}).
\newblock
\urldef\tempurl%
\url{https://doi.org/10.48550/ARXIV.2012.15029}
\showDOI{\tempurl}


\bibitem[\protect\citeauthoryear{of~Health}{of~Health}{2023}]%
        {allofus}
\bibfield{author}{\bibinfo{person}{National~Institutes of Health}.}
  \bibinfo{year}{2023}\natexlab{}.
\newblock \bibinfo{title}{All of Us}.
\newblock
\newblock
\newblock
\shownote{https://allofus.nih.gov.}


\bibitem[\protect\citeauthoryear{Park}{Park}{2022}]%
        {rcdm}
\bibfield{author}{\bibinfo{person}{et~al. Park, C.}}
  \bibinfo{year}{2022}\natexlab{}.
\newblock \showarticletitle{Development and Validation of the Radiology Common
  Data Model (R-CDM) for the International Standardization of Medical Imaging
  Data}.
\newblock \bibinfo{journal}{\emph{Yonsei medical journal}}
  \bibinfo{volume}{63(Suppl)} (\bibinfo{year}{2022}),
  \bibinfo{pages}{S74–S83}.
\newblock
\urldef\tempurl%
\url{https://doi.org/10.3349/ymj.2022.63.S74}
\showURL{%
\tempurl}


\bibitem[\protect\citeauthoryear{Puppala}{Puppala}{2015}]%
        {meteor}
\bibfield{author}{\bibinfo{person}{et~al. Puppala, M.}}
  \bibinfo{year}{2015}\natexlab{}.
\newblock \showarticletitle{METEOR: An Enterprise Health Informatics
  Environment to Support Evidence-Based Medicine}.
\newblock \bibinfo{journal}{\emph{IEEE transactions on bio-medical
  engineering}} \bibinfo{volume}{62}, \bibinfo{number}{12}
  (\bibinfo{year}{2015}), \bibinfo{pages}{2776–2786}.
\newblock
\urldef\tempurl%
\url{https://doi.org/10.1109/TBME.2015.2450181}
\showDOI{\tempurl}


\bibitem[\protect\citeauthoryear{Rubin and Desser}{Rubin and Desser}{2008}]%
        {Rubin2008210}
\bibfield{author}{\bibinfo{person}{Daniel~L. Rubin} {and}
  \bibinfo{person}{Terry~S. Desser}.} \bibinfo{year}{2008}\natexlab{}.
\newblock \showarticletitle{A Data Warehouse for Integrating Radiologic and
  Pathologic Data}.
\newblock \bibinfo{journal}{\emph{Journal of the American College of
  Radiology}} \bibinfo{volume}{5}, \bibinfo{number}{3} (\bibinfo{year}{2008}),
  \bibinfo{pages}{210 – 217}.
\newblock


\bibitem[\protect\citeauthoryear{Wang, Peng, Lu, Lu, Bagheri, and Summers}{Wang
  et~al\mbox{.}}{2017}]%
        {Wang20173462}
\bibfield{author}{\bibinfo{person}{Xiaosong Wang}, \bibinfo{person}{Yifan
  Peng}, \bibinfo{person}{Le Lu}, \bibinfo{person}{Zhiyong Lu},
  \bibinfo{person}{Mohammadhadi Bagheri}, {and} \bibinfo{person}{Ronald~M.
  Summers}.} \bibinfo{year}{2017}\natexlab{}.
\newblock \showarticletitle{ChestX-ray8: Hospital-scale chest X-ray database
  and benchmarks on weakly-supervised classification and localization of common
  thorax diseases}. In \bibinfo{booktitle}{\emph{30th IEEE Conference on
  Computer Vision and Pattern Recognition}} \emph{(\bibinfo{series}{CVPR
  2017})}, Vol.~\bibinfo{volume}{2017-January}. \bibinfo{pages}{3462 – 3471}.
\newblock
\urldef\tempurl%
\url{https://doi.org/10.1109/CVPR.2017.369}
\showDOI{\tempurl}


\end{thebibliography}

\end{document}